\pdfoutput=1

\documentclass[11pt]{article}

\usepackage[final]{acl}

\usepackage{times}
\usepackage{latexsym}

\usepackage[T1]{fontenc}

\usepackage[utf8]{inputenc}

\usepackage{microtype}

\usepackage{inconsolata}

\usepackage{graphicx}
\usepackage{booktabs}
\usepackage{multirow}
\usepackage{adjustbox}
\usepackage{pgfplots}
\usepackage{pgfplotstable}
\pgfplotsset{compat=1.18}
\usepgfplotslibrary{groupplots}
\usepackage{caption}
\DeclareCaptionFont{helv}{\sf\small}

\definecolor{UniBlau}{cmyk}{1,0.7,0,0}
\definecolor{UniGruen}{cmyk}{0.6,0,1,0}
\definecolor{UniOrange}{cmyk}{0,0.3,1,0}
\definecolor{UniRot}{cmyk}{0.4,1,0,0}

\pgfplotscreateplotcyclelist{mycyclelist}{
{UniBlau, mark=o},
{UniGruen,mark=square},
{UniOrange,mark=diamond},
{UniRot,mark=triangle},
{UniBlau!50!UniGruen,mark=*},
{UniGruen!50!UniOrange,mark=square*},
{UniOrange!50!UniRot,mark=triangle*},
{UniRot!50!UniBlau,mark=diamond*},
}

\pgfplotsset{every axis/.append style={
                    xlabel={$x$},          
                    ylabel={$y$},          
                    label style={font=\sffamily},
                    tick label style={font=\sffamily\scriptsize},
                    xticklabel style = {font=\sffamily\scriptsize},
                    title style = {font=\normalsize\sffamily},
                    ylabel near ticks,
                    y label style={font=\sffamily\scriptsize},
                    xlabel near ticks,
                    x label style={font=\sffamily\scriptsize},
                    legend cell align={left},
                    legend style={draw=none, font=\sffamily\scriptsize},  
                    },
                    }

\title{Learning from Litigation: Graphs for Retrieval and Reasoning in eDiscovery}

\author{
 \textbf{Sounak Lahiri\textsuperscript{1 $\ast$}},
 \textbf{Sumit Pai\textsuperscript{2 $\ast$}},
 \textbf{Tim Weninger\textsuperscript{3 $\star$}},
 \textbf{Sanmitra Bhattacharya\textsuperscript{4 $\dag$}},
\\
 \textsuperscript{$\ast$}Deloitte \& Touche Assurance \& Enterprise Risk Services India Private Limited, \\Bangalore, India \\
 \textsuperscript{$\dag$}Deloitte \& Touche LLP, New York City, USA \\
 \textsuperscript{$\star$}University of Notre Dame, Notre Dame, USA
\\
 \small{
   \textsuperscript{1}\href{solahiri@deloitte.com}{solahiri@deloitte.com}, 
   \textsuperscript{2}\href{sumpai@deloitte.com}{sumpai@deloitte.com},
   \textsuperscript{3}\href{tweninger@nd.edu}{tweninger@nd.edu},
   \textsuperscript{4}\href{sanmbhattacharya@deloitte.com}{sanmbhattacharya@deloitte.com}
 }
}

\begin{document}
\maketitle
\begin{abstract}
Electronic Discovery (eDiscovery) requires identifying relevant documents from vast collections for legal production requests. While artificial intelligence (AI) and natural language processing (NLP) have improved document review efficiency, current methods still struggle with legal entities, citations, and complex legal artifacts. To address these challenges, we introduce DISCOvery Graph (DISCOG), an \emph{emerging} system that integrates knowledge graphs for enhanced document ranking and classification, augmented by LLM-driven reasoning. DISCOG outperforms strong baselines in F1-score, precision, and recall across both balanced and imbalanced datasets. In real-world deployments, it has reduced litigation-related document review costs by approximately 98\%, demonstrating significant business impact.
\end{abstract}

\section{Introduction}


\begin{figure*}[!hbtp]
    \centering
    \includegraphics[width=\linewidth]{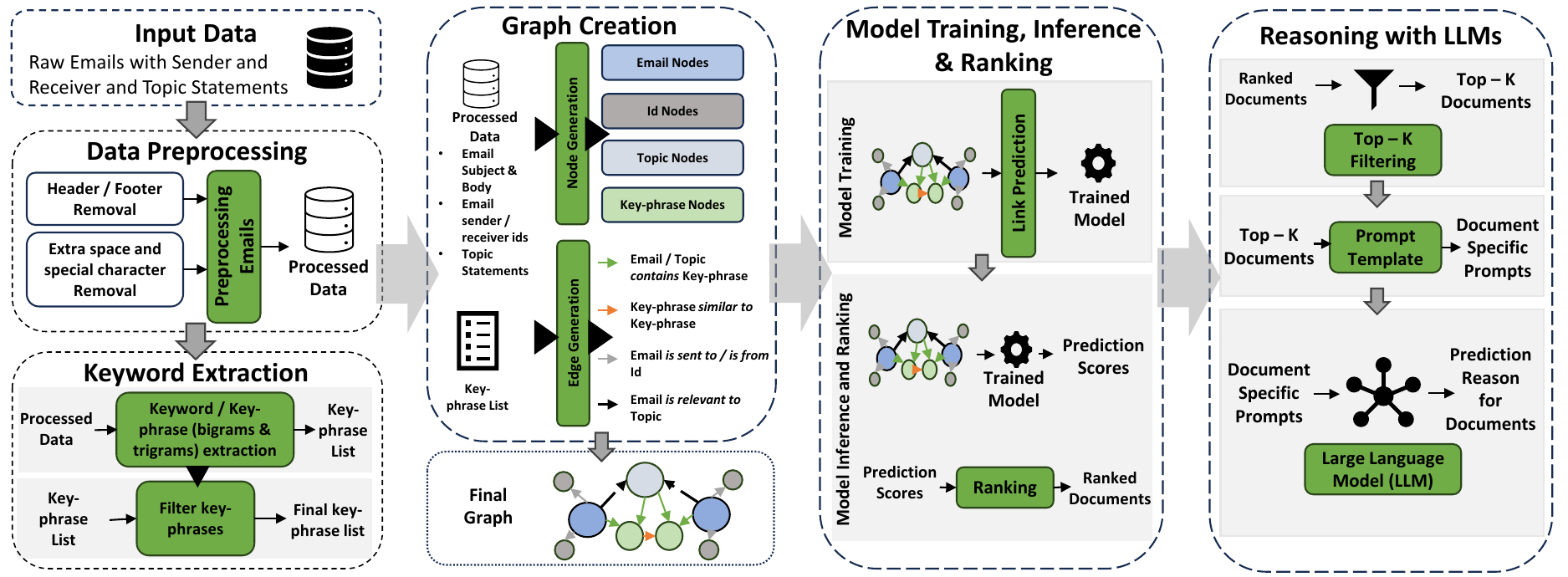}
    \caption{DISCOG: A heterogeneous graph-based approach for predictive coding and ranking in eDiscovery}
    \label{fig:graph_solution}
\end{figure*}

During legal proceedings, such as investigations, regulatory reviews, and litigation, parties engage in a legal process called \emph{discovery}, formally requesting relevant documents from opposing parties. Traditionally, this involves manually sifting through vast document repositories, a slow and costly process prone to human error. Electronic discovery (eDiscovery) encompasses the collection, review, and organization of digital documents, such as emails, contracts, and articles, to identify those relevant to discovery requests. Technology-assisted review (TAR) typically involves iterative workflows in which skilled professionals annotate documents for relevance guiding supervised learning models in prioritizing documents for review. Early TAR workflows relied on Boolean text queries but have since evolved to incorporate ranked retrieval, relevance feedback, and active learning techniques~\cite{sansone2022legal}. Recently, \emph{predictive coding}, which trains binary text classifiers to determine whether a document is relevant to a production request, has gained widespread use~\cite{brown2015peeking}. Large Language Models (LLMs) have also been explored for document relevance classification in eDiscovery~\cite{pai-etal-2023-exploration}. However, these text-only models struggle to effectively capture entities, citations, and other complex legal information frequently found in legal production requests, limiting their adoption. To address these challenges, we introduce DISCOvery Graph ({\bf DISCOG}), a novel \emph{emerging} approach that constructs a knowledge graph from the complex structural information within document corpus and leverages it to enhance document classification and ranking.

DISCOG frames the eDiscovery problem as a link prediction task within a knowledge graph, augmented by a Large Language Model (LLM) for reasoning. The graph consists of documents (\textit{e.g.}, email subjects and bodies from the EDRM corpus), topic statements, senders, and receivers. Keywords and keyphrases extracted from documents serve as additional nodes, with semantically similar keywords linked to enhance structural richness. Document relevance is determined through link prediction between document and topic nodes, where a document is classified as relevant if a link exists between them. To model the knowledge graph, DISCOG employs representation learning techniques, including Knowledge Graph Embedding (KGE) methods such as TransE~\cite{transE} and ComplEx~\cite{complEx}, as well as Graph Neural Networks (GNNs) like GraphSAGE~\cite{graphsage}. The trained model ranks documents by prediction probability, selecting the top $K$ documents (determined by a predefined recall threshold, typically 80\% \cite{halskov2013}) for further reasoning via LLMs. Fig.~\ref{fig:graph_solution} provides an overview of the DISCOG framework.

Due to the confidentiality of legal discovery processes, direct experimentation on real-world litigation is not feasible. Instead, we evaluate DISCOG using the publicly available Electronic Discovery Reference Model (EDRM) Enron Emails Dataset, previously used in the Text Retrieval Conference (TREC) Legal Track (2009–2011)\footnote{\url{https://trec-legal.umiacs.umd.edu/}} \cite{hedin2009overview, Grossman2011OverviewOT}. This dataset, which includes production requests and human-labeled relevance judgments, remains a benchmark for NLP research on LLM applications \cite{10.14778/3681954.3681994, chen-etal-2024-learnable, huang-etal-2024-large} and graph-based methods~\cite{shakiba2023correlationclusteringalgorithmdynamic, nouranizadeh2024contrastiverepresentationlearningdynamic}. By demonstrating DISCOG’s effectiveness on this established benchmark, we showcase its potential for real-world eDiscovery tasks.

\section{Related Work}



Prior research on the EDRM Enron dataset has primarily employed traditional information retrieval (IR) techniques \cite{Grossman2011OverviewOT,bm25}, where queries are executed against a document index to generate ranked lists of relevant documents. 



Transformer-based architectures \cite{vaswani2017attention} have transformed NLP by enabling cross-domain knowledge transfer with limited training data \cite{raffel2020exploring}. Models such as Contextualized Late Interaction over BERT (ColBERT) \cite{khattab2020colbert} and its improved variant, ColBERT v2 \cite{colbert}, leverage contextualized embeddings and late interaction mechanisms to enhance document ranking. For adaptation to the legal domain, \cite{goldilocks} pre-trained BERT on legal data and fine-tuning based on human review for active learning. Recently, large lanuage models (LLMs) have been used in several use-cases for identify relevancy based on semantic relations and generating responses along with appropriate reasoning. \cite{pai-etal-2023-exploration} experimented with out of the box and fine-tuned LLMs for classification of documents relevant to a topic. \cite{combining_llms_with_TAR} additionally proposed active learning methods to rank the classifications obtained from LLMs. However, despite their strength in text processing, these models often overlook relational dependencies crucial in legal contexts.

To address this limitation, legal data can be structured as graphs, where documents, topics, and entities form nodes, and relationships define edges. Graph-based methods have been widely applied in areas such as social networks and biomedical research, offering structured representations of interconnected data \cite{kg}. Graph representation learning captures latent semantic relationships by embedding nodes and edges into low-dimensional spaces, optimizing them for classification and link prediction tasks. \cite{CaseGNN} proposed a text-attributed case graph (TACG) with downstream applications using graph attention trained with contrastive learning methods. \cite{CaseGNN++} builds on top of \cite{CaseGNN} with updated attention layer to deal with both nodes and edges and graph augmentation technique for better learning. \cite{caselink} creates a Global Case Graph and employs inductive graph learning for various use-cases. \cite{qabisar} and \cite{finding_the_law} provides similar approaches for Statutory Articles.

Two main approaches dominate graph representation learning: (1) Knowledge Graph Embedding (KGE) models and (2) Graph Neural Networks (GNNs). KGE models, including TransE \cite{transE}, ComplEx \cite{complEx}, RotatE \cite{sun2019rotate}, and DistMult \cite{yang2015embedding}, generate embeddings through lookup tables and optimize them using scoring functions. GNNs, in contrast, aggregate node features from their neighborhoods over multiple hops ($n$-hops), enabling more expressive representations \cite{zhou2021graph}.

Among GNNs, GraphSAGE \cite{graphsage} constructs node embeddings by aggregating sampled neighbor information, while Graph Attention Networks (GAT) \cite{veličković2018graph} enhance this by assigning attention scores to different neighbors. Relation Graph Convolutional Networks (RGCNs) \cite{schlichtkrull2017modeling} further extend GCNs by incorporating different edge types, making them well-suited for heterogeneous legal data. These graph-based approaches provide a structured way to model complex dependencies in legal discovery, addressing limitations of purely text-based methods.

\section{Methodology}

\begin{figure}
    \centering
    \includegraphics[width=\linewidth]{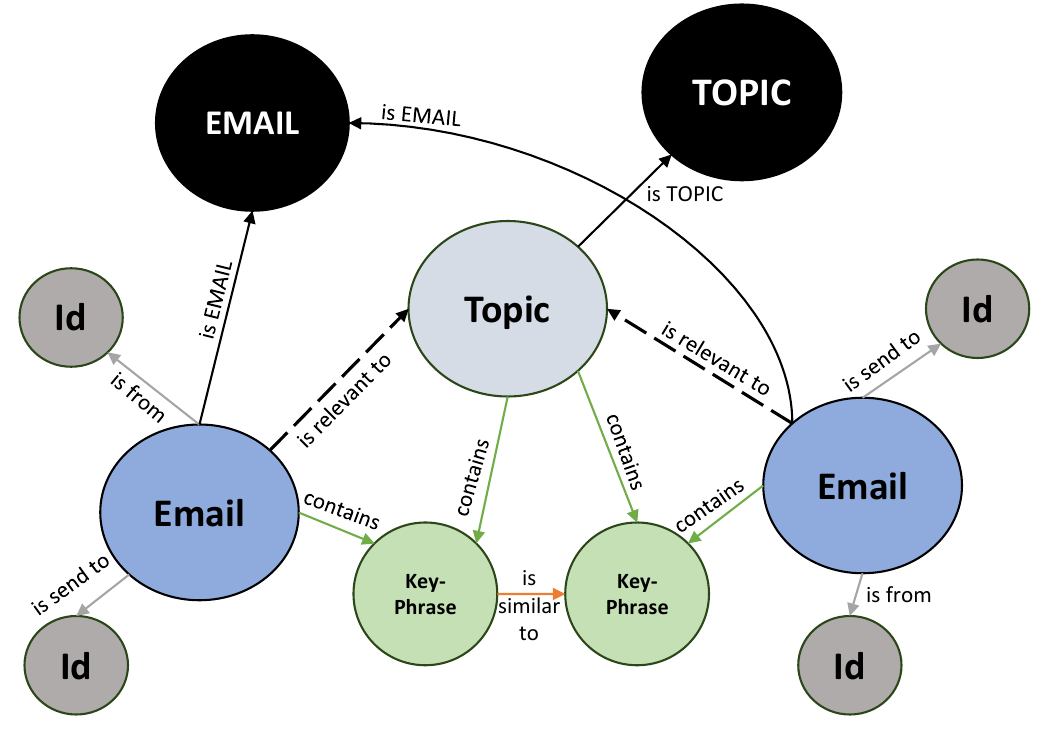}    
    \vspace{-.3cm}
    \caption{Graph Structure - Schematic Diagram}
    \label{fig:graph_structure}
\end{figure}

This study tackles predictive coding in eDiscovery by constructing a heterogeneous knowledge graph from documents, emails, topic statements, and metadata (\textit{e.g.}, email IDs). Semantic relationships are derived from keywords and keyphrases, and link prediction techniques classify document relevance by predicting links between document and topic nodes. We employ both Knowledge Graph Embedding (KGE) methods (\textit{e.g.}, TransE, ComplEx) and Graph Neural Networks (GNNs) (\textit{e.g.}, GraphSAGE). While KGE methods learn low-dimensional node and edge embeddings, GNNs aggregate features from neighboring nodes to enhance link prediction accuracy. The trained models rank documents by relevance, with an LLM providing reasoning for predictions—addressing both classification and interpretability in legal document review.



\subsection{Dataset}

The EDRM Enron Emails dataset, used in the TREC Legal Track (2009–2011), contains 455,449 emails and 230,143 attachments \cite{Grossman2011OverviewOT}. As a case study, we focus on production requests from the 2009 and 2011 tracks, covering seven topics in 2009 and three in 2011. Each topic includes a seed document set for training and \textit{qrels}, which provide human-assessed relevance judgments for evaluation. The full topic details and data distribution are provided in Appendix~\ref{dataset_appendix}.


\begin{figure*}[t]
    \centering
        \begin{tikzpicture}
    \begin{groupplot}[
        group style={group name=mygroup, group size=1 by 3, vertical sep=.2cm, xlabels at=edge bottom, xticklabels at=edge bottom,},
        ybar,
        /pgf/bar width=3pt,
        cycle list name = mycyclelist,
        every axis plot/.append style={fill},
        no marks,
        symbolic x coords={ColBERT v2, TransE, ComplEx, GAT, RGCN (TransE), GraphSAGE},
        xtick=data,
        ymin=0, ymax=1,
        enlarge x limits=0.1,
        width=1.04\textwidth,
        height=3cm,
        title style={align=center},
        xlabel={Models},
        ylabel={F1},
    ]
    \nextgroupplot[
        ylabel={F1},
    ]    
        \addplot coordinates {(ColBERT v2, 0.57) (TransE, 0.71) (ComplEx, 0.71) (GAT, 0.65) (RGCN (TransE), 0.77) (GraphSAGE, 0.83)};
        \addplot coordinates {(ColBERT v2, 0.75) (TransE, 0.83) (ComplEx, 0.82) (GAT, 0.71) (RGCN (TransE), 0.80) (GraphSAGE, 0.86)};
        \addplot coordinates {(ColBERT v2, 0.49) (TransE, 0.73) (ComplEx, 0.77) (GAT, 0.64) (RGCN (TransE), 0.61) (GraphSAGE, 0.86)};
        \addplot coordinates {(ColBERT v2, 0.62) (TransE, 0.58) (ComplEx, 0.59) (GAT, 0.54) (RGCN (TransE), 0.53) (GraphSAGE, 0.82)};
        \addplot coordinates {(ColBERT v2, 0.74) (TransE, 0.65) (ComplEx, 0.82) (GAT, 0.51) (RGCN (TransE), 0.49) (GraphSAGE, 0.84)};
        \addplot coordinates {(ColBERT v2, 0.48) (TransE, 0.61) (ComplEx, 0.65) (GAT, 0.52) (RGCN (TransE), 0.56) (GraphSAGE, 0.54)};
        \addplot coordinates {(ColBERT v2, 0.56) (TransE, 0.59) (ComplEx, 0.68) (GAT, 0.69) (RGCN (TransE), 0.84) (GraphSAGE, 0.88)};
        \addplot coordinates {(ColBERT v2, 0.68) (TransE, 0.73) (ComplEx, 0.83) (GAT, 0.71) (RGCN (TransE), 0.76) (GraphSAGE, 0.90)};
        \addplot coordinates {(ColBERT v2, 0.62) (TransE, 0.70) (ComplEx, 0.79) (GAT, 0.75) (RGCN (TransE), 0.59) (GraphSAGE, 0.89)};
        \addplot coordinates {(ColBERT v2, 0.56) (TransE, 0.67) (ComplEx, 0.82) (GAT, 0.65) (RGCN (TransE), 0.53) (GraphSAGE, 0.85)};
    \nextgroupplot[
        ylabel={Precision},
    ]
        \addplot coordinates {(ColBERT v2, 0.81) (TransE, 0.70) (ComplEx, 0.70) (GAT, 0.71) (RGCN (TransE), 0.76) (GraphSAGE, 0.80)};
        \addplot coordinates {(ColBERT v2, 0.90) (TransE, 0.88) (ComplEx, 0.91) (GAT, 0.76) (RGCN (TransE), 0.86) (GraphSAGE, 0.83)};
        \addplot coordinates {(ColBERT v2, 0.86) (TransE, 0.70) (ComplEx, 0.84) (GAT, 0.73) (RGCN (TransE), 0.60) (GraphSAGE, 0.83)};
        \addplot coordinates {(ColBERT v2, 0.84) (TransE, 0.58) (ComplEx, 0.60) (GAT, 0.55) (RGCN (TransE), 0.53) (GraphSAGE, 0.80)};
        \addplot coordinates {(ColBERT v2, 0.88) (TransE, 0.80) (ComplEx, 0.81) (GAT, 0.51) (RGCN (TransE), 0.54) (GraphSAGE, 0.83)};
        \addplot coordinates {(ColBERT v2, 0.46) (TransE, 0.61) (ComplEx, 0.63) (GAT, 0.52) (RGCN (TransE), 0.56) (GraphSAGE, 0.54)};
        \addplot coordinates {(ColBERT v2, 0.78) (TransE, 0.60) (ComplEx, 0.68) (GAT, 0.72) (RGCN (TransE), 0.83) (GraphSAGE, 0.88)};
        \addplot coordinates {(ColBERT v2, 0.68) (TransE, 0.76) (ComplEx, 0.84) (GAT, 0.76) (RGCN (TransE), 0.76) (GraphSAGE, 0.90)};
        \addplot coordinates {(ColBERT v2, 0.90) (TransE, 0.79) (ComplEx, 0.79) (GAT, 0.79) (RGCN (TransE), 0.68) (GraphSAGE, 0.91)};
        \addplot coordinates {(ColBERT v2, 0.87) (TransE, 0.70) (ComplEx, 0.92) (GAT, 0.70) (RGCN (TransE), 0.53) (GraphSAGE, 0.85)};
    \nextgroupplot[
        ylabel={Recall},
    ]
        \addplot coordinates {(ColBERT v2, 0.55) (TransE, 0.73) (ComplEx, 0.72) (GAT, 0.63) (RGCN (TransE), 0.78) (GraphSAGE, 0.88)};
        \addplot coordinates {(ColBERT v2, 0.69) (TransE, 0.80) (ComplEx, 0.77) (GAT, 0.68) (RGCN (TransE), 0.80) (GraphSAGE, 0.83)};
        \addplot coordinates {(ColBERT v2, 0.51) (TransE, 0.82) (ComplEx, 0.75) (GAT, 0.61) (RGCN (TransE), 0.63) (GraphSAGE, 0.86)};
        \addplot coordinates {(ColBERT v2, 0.60) (TransE, 0.58) (ComplEx, 0.59) (GAT, 0.54) (RGCN (TransE), 0.53) (GraphSAGE, 0.84)};
        \addplot coordinates {(ColBERT v2, 0.69) (TransE, 0.63) (ComplEx, 0.83) (GAT, 0.51) (RGCN (TransE), 0.51) (GraphSAGE, 0.84)};
        \addplot coordinates {(ColBERT v2, 0.46) (TransE, 0.58) (ComplEx, 0.68) (GAT, 0.50) (RGCN (TransE), 0.51) (GraphSAGE, 0.49)};
        \addplot coordinates {(ColBERT v2, 0.56) (TransE, 0.58) (ComplEx, 0.66) (GAT, 0.67) (RGCN (TransE), 0.75) (GraphSAGE, 0.89)};
        \addplot coordinates {(ColBERT v2, 0.66) (TransE, 0.70) (ComplEx, 0.79) (GAT, 0.75) (RGCN (TransE), 0.75) (GraphSAGE, 0.92)};
        \addplot coordinates {(ColBERT v2, 0.56) (TransE, 0.60) (ComplEx, 0.75) (GAT, 0.71) (RGCN (TransE), 0.58) (GraphSAGE, 0.83)};
        \addplot coordinates {(ColBERT v2, 0.56) (TransE, 0.68) (ComplEx, 0.75) (GAT, 0.60) (RGCN (TransE), 0.56) (GraphSAGE, 0.83)};
    \end{groupplot}
    \end{tikzpicture}
    \vspace{-.7cm}
    \caption{Predictive coding performance of baselines and graph-based models. Grouped bars represent Topics 201--403 in numeric order left to right; their specific identification is relevant this illustration.}
\label{tab:predictive_coding_results}
\end{figure*}

\subsection{Baselines for Predictive Coding}

We evaluate DISCOG against two widely used predictive coding baselines in eDiscovery:  

\textbf{BM25L}: A standard IR model that ranks documents based on query relevance~\cite{bm25l}. In our setup, the topic statement serves as the query, and BM25L computes a relevance score for each document based on term frequency, document length, and other factors.  

\textbf{ColBERT v2}: A Transformer-based retrieval model optimized for passage ranking. We use a pretrained ColBERT v2 model with frozen weights and a downstream classifier to refine relevance predictions, leveraging ColBERT’s contextualized embeddings for improved predictive coding.  

\subsection{DISCOvery Graph (DISCOG)}

DISCOG employs a graph-based predictive coding approach in three stages: (1) it constructs a heterogeneous knowledge graph from extracted keywords, documents, topics, senders, and receivers; (2) it applies predictive coding using KGE methods (TransE, ComplEx) and GNN models (GraphSAGE, GAT, RGCN) to learn relationships and improve classification accuracy; (3) the trained model ranks documents by predicted relevance to topic statements, capturing complex relational dependencies to enhance predictive performance.  

\subsubsection{{Graph Construction}}
\label{graph_creation_method}

To harness relational structures in text data, we construct a heterogeneous knowledge graph consisting of four node types: \textit{documents/emails}, \textit{senders/recipients}, \textit{topics}, and \textbf{keywords/keyphrases}. Keywords/keyphrases are a combination of unigrams, bigrams and trigrams and extracted from documents using the subject and body and from topic statements using KeyBERT \cite{grootendorst2020keybert} and used as distinct nodes. To reduce noise, we retain only keywords appearing in at least five documents. These are connected to one another based on semantic similarity obtained by a cosine similarity score of 0.75 and above.


Most knowledge graph embedding methods are transductive, making inference on unseen nodes challenging \cite{luca_costabello_2023_7818821}. To address this, we introduce two master nodes: \textbf{DOCUMENT} and \textbf{TOPIC}, linking all documents and topics to their respective master nodes. The master node \textbf{DOCUMENT} is connected to all nodes obtained from documents and ensures that no isolated nodes are present during inference. Similar connections are followed for topic nodes. These master nodes are only used during transductive embedding generation and are unnecessary for graph neural networks, which are inductive and handle unseen nodes inherently. A schematic diagram of the graph is shown in Figure \ref{fig:graph_structure}

The graph incorporates links from the seed and qrels sets. For knowledge graph embedding, only positive links---indicating document relevance---are included to align with the open-world assumption \cite{luca_costabello_2023_7818821}. In contrast, graph neural networks leverage both positive and negative links, improving their ability to distinguish relevant from non-relevant documents.

\subsubsection{{Predictive Coding}}

DISCOG formulates predictive coding as a link prediction task within a knowledge graph. Two modeling approaches are employed: Knowledge Graph Embeddings (KGE) and Graph Neural Networks (GNNs).  

For \textbf{KGE-based prediction}, TransE and ComplEx learn low-dimensional node embeddings by minimizing triplet loss with multi-class negative log-likelihood \cite{ampligraph}. Training considers only relevant links from the seed set, represented as triples $\langle Document_{i}, relevant\_to, Topic_{j} \rangle$. During inference, confidence scores for predicted links are calibrated using ground-truth labels, with a classification threshold optimized for F1-score on the validation set.  

For \textbf{GNN-based prediction}, node embeddings are initialized using Sentence Transformers and refined via GraphSAGE, GAT, and RGCN, integrated with TransE. Unlike KGE, GNN training incorporates both relevant and non-relevant links, assigning edge values of 1 (relevant) and 0 (non-relevant). A classification head predicts edge labels, and edge scores are thresholded to optimize macro average F1-score during inference.  

Both approaches enable DISCOG to classify documents as relevant or non-relevant, leveraging graph structure to enhance predictive coding in eDiscovery.

\subsubsection{{Ranking and LLM Prediction}}

Documents are ranked based on edge scores, normalized via min-max scaling for KGE methods, while GNNs use classification probabilities directly. Performance is evaluated using Recall@$k$, and results are benchmarked against BM25L and BERT with a classifier.

Finally, building on \citet{pai-etal-2023-exploration}, we apply LLMs to explain predictions. The top $K$ ranked documents are selected, and GPT-3.5-turbo is queried Out-Of-Box (OOB) with a prompt designed to validate graph model predictions and generate reasoning. The prompt is upgraded from the work in \citet{pai-etal-2023-exploration}, to incorporate the prediction results from the GNN or KGE based method, along with the keywords identified from the document, and the overall LLM task is modified to validate the model's prediction along with a reason to support its decision.

\section{Results}

We evaluate DISCOG using emails from the EDRM Enron Dataset, excluding attachments. This section details the graph construction, predictive coding and ranking outcomes, and an analysis of cost savings and business impact.  

\subsection{Heterogeneous Information Network}

The final graph consists of 455,449 email nodes, ten topic nodes, and 34,134 keyword nodes extracted from emails and topic statements. Additionally, 103,926 distinct sender/receiver IDs are included. Edges are formed based on email-to-keyword associations, with similar keywords linked. The number of \textbf{Emails} \emph{relevant to} \textbf{Topic} edges varies per topic, determined by the seed and qrels sets used for model training and evaluation.  

\begin{figure}[t]
    \centering
    \input{images/recall}
    \vspace{-.3cm}
    \caption{Recall@k plots for topics at different values of $k$ ranging from 0 to total count of emails in the dataset.}
    \label{fig:recall@k}
\end{figure}

\begin{figure}[!th]
    \centering
    \pgfplotstableread{
x  twothou fivethou twnty fifty hundred twohundred
1 0.05 0.05 0.16 0.22 0.29 0.29
2 0.19 0.34 0.38 0.4 0.35 0.27
3 0.31 0.37 0.42 0.38 0.3 0.27
4 0.27 0.33 0.42 0.36 0.31 0.24
5 0.36 0.36 0.35 0.29 0.24 0.22
6 0.44 0.54 0.58 0.52 0.45 0.32
7 0.68 0.73 0.65 0.52 0.41 0.29
}{\twozeroone}

\pgfplotstableread{
x  twothou fivethou twnty fifty hundred twohundred
1 nan 0.02 0.08 0.1 0.16 0.13
2 0.32 0.31 0.4 0.3 0.24 0.16
3 0.26 0.3 0.23 0.19 0.16 0.11
4 0.15 0.16 0.23 0.19 0.17 0.13
5 0.37 0.38 0.26 0.15 0.14 0.11
6 0.0 0.07 0.18 0.22 0.25 0.16
7 0.52 0.53 0.54 0.38 0.24 0.15
}{\twozerotwo}

\pgfplotstableread{
x  twothou fivethou twnty fifty hundred twohundred
1 0.47 0.45 0.43 0.54 0.48 0.4
2 0.67 0.72 0.64 0.54 0.46 0.37
3 0.45 0.51 0.57 0.54 0.48 0.37
4 0.53 0.61 0.6 0.52 0.46 0.37
5 0.12 0.23 0.32 0.26 0.21 0.18
6 0.02 0.04 0.15 0.23 0.32 0.27
7 0.73 0.72 0.63 0.55 0.45 0.37
}{\twozerothree}

\pgfplotstableread{
x  twothou fivethou twnty fifty hundred twohundred
1 0.13 0.16 0.23 0.25 0.34 0.37
2 0.49 0.62 0.66 0.62 0.51 0.41
3 0.07 0.11 0.2 0.28 0.3 0.34
4 0.09 0.15 0.23 0.29 0.32 0.33
5 0.02 0.02 0.05 0.1 0.16 0.23
6 0.06 0.06 0.12 0.21 0.25 0.28
7 0.62 0.69 0.69 0.6 0.52 0.42
}{\twozerofour}

\pgfplotstableread{
x  twothou fivethou twnty fifty hundred twohundred
1 0.34 0.42 0.62 0.7 0.68 0.58
2 0.48 0.62 0.75 0.74 0.63 0.51
3 0.28 0.34 0.34 0.33 0.29 0.27
4 0.42 0.56 0.66 0.67 0.59 0.46
5 0.02 0.04 0.06 0.13 0.18 0.25
6 0.0 0.02 0.07 0.13 0.19 0.23
7 0.39 0.57 0.71 0.69 0.57 0.46
}{\twozerofive}

\pgfplotstableread{
x  twothou fivethou twnty fifty hundred twohundred
1 0.39 0.49 0.43 0.32 0.27 0.21
2 nan 0.03 0.2 0.17 0.19 0.2
3 0.1 0.15 0.27 0.26 0.21 0.17
4 0.33 0.37 0.37 0.32 0.29 0.23
5 nan 0.03 0.09 0.13 0.14 0.15
6 0.11 0.12 0.16 0.2 0.21 0.17
7 0.04 0.07 0.09 0.11 0.15 0.14
}{\twozerosix}

\pgfplotstableread{
x  twothou fivethou twnty fifty hundred twohundred
1 0.42 0.62 0.72 0.72 0.72 0.69
2 0.59 0.77 0.86 0.82 0.78 0.7
3 0.04 0.09 0.22 0.36 0.47 0.54
4 0.13 0.22 0.44 0.55 0.61 0.64
5 0.4 0.48 0.59 0.59 0.58 0.57
6 0.45 0.62 0.78 0.79 0.78 0.71
7 0.65 0.84 0.86 0.83 0.79 0.72
}{\twozeroseven}

\pgfplotstableread{
x  twothou fivethou twnty fifty hundred twohundred
1 0.06 0.06 0.2 0.37 0.48 0.56
2 0.16 0.28 0.54 0.68 0.75 0.69
3 0.36 0.47 0.54 0.55 0.52 0.45
4 0.26 0.41 0.6 0.68 0.66 0.62
5 0.11 0.21 0.4 0.49 0.51 0.52
6 0.04 0.09 0.28 0.46 0.65 0.67
7 0.28 0.43 0.71 0.79 0.78 0.69
}{\fourzeroone}

\pgfplotstableread{
x  twothou fivethou twnty fifty hundred twohundred
1 0.02 0.02 0.12 0.29 0.46 0.41
2 0.51 0.65 0.72 0.64 0.54 0.4
3 0.39 0.44 0.46 0.39 0.32 0.27
4 0.33 0.51 0.63 0.61 0.52 0.4
5 0.25 0.32 0.48 0.46 0.4 0.35
6 0.06 0.09 0.2 0.2 0.22 0.32
7 0.64 0.74 0.74 0.66 0.55 0.41
}{\fourzerotwo}

\pgfplotstableread{
x  twothou fivethou twnty fifty hundred twohundred
1 0.14 0.19 0.26 0.24 0.22 0.22
2 0.63 0.6 0.46 0.33 0.25 0.17
3 0.33 0.34 0.28 0.2 0.15 0.11
4 0.26 0.35 0.33 0.28 0.21 0.15
5 0.14 0.17 0.14 0.12 0.12 0.11
6 nan 0.07 0.09 0.1 0.12 0.13
7 0.63 0.55 0.45 0.32 0.23 0.17
}{\fourzerothree}

\pgfplotsset{every axis title/.append style={at={(0.08,0.48)}, font=\footnotesize}}

\begin{tikzpicture}
\centering
    \begin{groupplot}[
        group style={group size=1 by 10,
            horizontal sep = .1 cm, 
            vertical sep = .1 cm, 
            ylabels at=edge left,
            xticklabels at=edge bottom,
            yticklabels at=edge left,
            xlabels at=edge bottom}, 
        xlabel = {},
        xticklabel style={align=center},
        xticklabels = {BM25L, ColBERT, TransE, ComplEx, GAT, RCGN\\(TransE), Graph\\SAGE},
        xtick={1,2,3,4,5,6,7},
        ymin=0,
        ymax=1.1,
        ybar=1pt,
        /pgf/bar width=2pt,
        ylabel = {F1},
        width = 8.3 cm, 
        height = 2.8 cm,
        legend columns=6, 
        legend style={at={(3,-12.8)}, anchor=north}, 
        every axis plot/.append style={fill},
        no marks,
        cycle list name = mycyclelist,
        ]
        \nextgroupplot[title={{\footnotesize 201}}, ]
        \addplot+ table [x=x, y=twothou] {\twozeroone};        
        \addplot+ table [x=x, y=fivethou] {\twozeroone};        
        \addplot+ table [x=x, y=twnty] {\twozeroone};
        \addplot+ table [x=x, y=fifty] {\twozeroone};
        \addplot+ table [x=x, y=hundred] {\twozeroone};
        \addplot+ table [x=x, y=twohundred] {\twozeroone};
        
        \nextgroupplot[title={{\footnotesize 202}}]
        \addplot+ table [x=x, y=twothou] {\twozerotwo};
        \addplot+ table [x=x, y=fivethou] {\twozerotwo};
        \addplot+ table [x=x, y=twnty] {\twozerotwo};
        \addplot+ table [x=x, y=fifty] {\twozerotwo};
        \addplot+ table [x=x, y=hundred] {\twozerotwo};
        \addplot+ table [x=x, y=twohundred] {\twozerotwo};
        
        \nextgroupplot[title={{\footnotesize 203}}, ylabel={F1}]
        \addplot+ table [x=x, y=twothou] {\twozerothree};
        \addplot+ table [x=x, y=fivethou] {\twozerothree};
        \addplot+ table [x=x, y=twnty] {\twozerothree};
        \addplot+ table [x=x, y=fifty] {\twozerothree};
        \addplot+ table [x=x, y=hundred] {\twozerothree};
        \addplot+ table [x=x, y=twohundred] {\twozerothree};
        
        \nextgroupplot[title={{\footnotesize 204}}]
        \addplot+ table [x=x, y=twothou] {\twozerofour};
        \addplot+ table [x=x, y=fivethou] {\twozerofour};
        \addplot+ table [x=x, y=twnty] {\twozerofour};
        \addplot+ table [x=x, y=fifty] {\twozerofour};
        \addplot+ table [x=x, y=hundred] {\twozerofour};
        \addplot+ table [x=x, y=twohundred] {\twozerofour};
        
        \nextgroupplot[title={{\footnotesize 205}}, ylabel={F1}]
        \addplot+ table [x=x, y=twothou] {\twozerofive};
        \addplot+ table [x=x, y=fivethou] {\twozerofive};
        \addplot+ table [x=x, y=twnty] {\twozerofive};
        \addplot+ table [x=x, y=fifty] {\twozerofive};
        \addplot+ table [x=x, y=hundred] {\twozerofive};
        \addplot+ table [x=x, y=twohundred] {\twozerofive};
        
        \nextgroupplot[title={{\footnotesize 206}}]
        \addplot+ table [x=x, y=twothou] {\twozerosix};
        \addplot+ table [x=x, y=fivethou] {\twozerosix};
        \addplot+ table [x=x, y=twnty] {\twozerosix};
        \addplot+ table [x=x, y=fifty] {\twozerosix};
        \addplot+ table [x=x, y=hundred] {\twozerosix};
        \addplot+ table [x=x, y=twohundred] {\twozerosix};    
        
        \nextgroupplot[title={{\footnotesize 207}}]
        \addplot+ table [x=x, y=twothou] {\twozeroseven};
        \addplot+ table [x=x, y=fivethou] {\twozeroseven};
        \addplot+ table [x=x, y=twnty] {\twozeroseven};
        \addplot+ table [x=x, y=fifty] {\twozeroseven};
        \addplot+ table [x=x, y=hundred] {\twozeroseven};
        \addplot+ table [x=x, y=twohundred] {\twozeroseven};
        
        \nextgroupplot[title={{\footnotesize 401}}]
        \addplot+ table [x=x, y=twothou] {\fourzeroone};
        \addplot+ table [x=x, y=fivethou] {\fourzeroone};
        \addplot+ table [x=x, y=twnty] {\fourzeroone};
        \addplot+ table [x=x, y=fifty] {\fourzeroone};
        \addplot+ table [x=x, y=hundred] {\fourzeroone};
        \addplot+ table [x=x, y=twohundred] {\fourzeroone};
        
        \nextgroupplot[title={{\footnotesize 402}}]
        \addplot+ table [x=x, y=twothou] {\fourzerotwo};
        \addplot+ table [x=x, y=fivethou] {\fourzerotwo};
        \addplot+ table [x=x, y=twnty] {\fourzerotwo};
        \addplot+ table [x=x, y=fifty] {\fourzerotwo};
        \addplot+ table [x=x, y=hundred] {\fourzerotwo};
        \addplot+ table [x=x, y=twohundred] {\fourzerotwo}; 

        
        \nextgroupplot[single ybar legend,legend to name=zelda, title={{\footnotesize 403}}] 
        \addplot+ table [x=x, y=twothou] {\fourzerothree};
        \addlegendentry{$k$=2000};
        \addplot+ table [x=x, y=fivethou] {\fourzerothree};
        \addlegendentry{$k$=5000};
        \addplot+ table [x=x, y=twnty] {\fourzerothree};
        \addlegendentry{$k$=20000};
        \addplot+ table [x=x, y=fifty] {\fourzerothree};
        \addlegendentry{$k$=50000};
        \addplot+ table [x=x, y=hundred] {\fourzerothree};
        \addlegendentry{$k$=100000};
        \addplot+ table [x=x, y=twohundred] {\fourzerothree};
        \addlegendentry{$k$=200000};

    \end{groupplot}   
    \ref{zelda}
\end{tikzpicture}
    \vspace{-.3cm}
    \caption{Ranking performance with prediction metrics as a function of $k$.}
    \label{fig:metrics_@k}
\end{figure}

\subsection{Predictive Coding Results}

We evaluate classification and ranking performance using qrels. Since BM25L is a ranking algorithm, it is excluded from classification evaluation. The classification results, summarized in Fig.~\ref{tab:predictive_coding_results}, show that the GNN-based GraphSAGE model consistently outperforms others, including RGCN and GAT. 

Most topics exhibit highly skewed distributions of relevant and non-relevant cases, leading to lower performance for baseline and KGE-based approaches. Despite this, GraphSAGE maintains strong performance across topics, with the exception of a single topic (\#206), which has the fewest relevant seed cases.

\subsection{Ranking Results}

Following the TREC 2009 and 2011 Legal Track evaluation scheme, we assess ranking performance using F1-score, precision, and recall at various cutoff values of $k$, where $k$ represents the number of reviewed documents. BM25L generates a natural ranking, while graph-based methods first classify documents before ranking them by confidence scores. Metrics are evaluated at thresholds: 2000 through 200000, as shown in Figs.~\ref{fig:recall@k} and \ref{fig:metrics_@k}.

GraphSAGE consistently outperforms other methods in ranking, achieving high recall across most topics. While knowledge graph models like RGCN and TransE show higher recall at lower $k$ values, overall, GraphSAGE demonstrates superior performance across both balanced and imbalanced datasets.

\begin{table*}[!t]
\centering
\caption{Reasoning results sample - Two examples have been shown for topic statements Online Trading and Derivative Trading (The predictions are taken from the Graph Model)}
\begin{tabular}{p{40mm}|p{37mm}|p{12mm}|p{55mm}}
\toprule
\textbf{Topic Statement} & \textbf{Document} & \textbf{Pred.} & \textbf{LLM Answer \& Reason} \\
\hline
\hline
\multirow{2}{40mm}{\footnotesize\textbf{All documents or communications that describe, discuss, refer to, report on, or relate to the design, development, operation, or marketing of enrononline, or any other online service offered, provided, or used by the Company (or any of its subsidiaries, predecessors, or successors-in-interest), for the purchase, sale, trading, or exchange of financial or other instruments or products, including but not limited to, derivative instruments, commodities, futures, and swaps.}} & 
\footnotesize \textbf{subject}: re: Aquila EOL trade assignments \newline The EOL trade assignment letters are prepared off the same form we do our regular assignments. there is no  special language in the assignment form about online issues.
& \footnotesize Relevant & \footnotesize Yes, the AI model is correct. The email discusses the use of a form for trade assignments, which is related to the trading of financial instruments and products, including derivatives and futures, on EnronOnline or other online services offered by Enron. Therefore, it is relevant to the topic identified by the keywords. \\
\cmidrule{2-4}
& \footnotesize \textbf{subject}: online trading \newline
per mark's conversation, you were going to get back to us on the scoop on France and utility counterparties.  any update? & \footnotesize Non-Relevant & \footnotesize Yes, the AI model is correct in its prediction that the email is not relevant to the given topic. The email is about a conversation regarding France and utility counterparties, and there is no mention of any online service offered, provided, or used by Enron for the purchase, sale, trading, or exchange of financial or other instruments or products. Therefore, it does not fall under the scope of the given topic.\\

\hline

\multirow{2}{40mm}{\footnotesize\textbf{All documents or communications that describe, discuss, refer to, report on, or relate to whether the purchase, sale, trading, or exchange of over-the-counter derivatives, or any other actual or contemplated financial instruments or products, is, was, would be, or will be legal or illegal, or permitted or prohibited, under any existing or proposed rule(s), regulation(s), law(s), standard(s), or other proscription(s), whether domestic or foreign.}} & \footnotesize \textbf{subject}: Credit Susie financial products (""csfb"") \newline
susan: csfb had a name change earlier this year.  can you please see that the blue file is changed?  also, please let the other paralegals know what to do in an instance such as this.  thanks.  sara & \footnotesize Non-Relevant & \footnotesize Yes, the AI model is correct in identifying the email as not relevant to the given topic. The email is a simple request to change a file related to a name change in Credit Suisse Financial Products and does not discuss or describe any purchase, sale, trading, or exchange of financial instruments or products, nor does it mention any legal or regulatory issues related to them. Therefore, it is not relevant to the given topic. \\
\cmidrule{2-4}
& \footnotesize \textbf{subject}: re: ECT Nymex hedge exemption applications
ECT has a policy that prohibits employees from trading in any commodities \newline
that ECT trades.  please make sure any future trading you may do complies with this policy.  thank you. & \footnotesize Relevant & \footnotesize No, the AI model is not correct. The email is not relevant to the given topic. The email only discusses the company policy on trading commodities and does not provide any information or discussion on the legality or permissibility of financial instruments or products. The keywords identified in the email are not directly related to the topic. \\ \bottomrule
\end{tabular}
\label{tab:reason}

\end{table*}

\subsection{LLM Reasoning Results}  

We use GPT-3.5 Turbo out-of-the-box (OOB) without instruction tuning to generate reasoning for predictions. Keywords extracted from documents guide the LLM, improving interpretability. Sample outputs are shown in Table~\ref{tab:reason}.  


Due to the augmentation of the Graph model's prediction result and the keywords identified from the documents, the LLM is better able to justify the reasoning behind the graph model's prediction. In most cases, the LLM agrees with the prediction and formulates a reason based on the observed keywords and its similarity with the topic statement. However, in instances where graph model misclassifies documents, the LLM can correct errors, as seen in the fourth example. This provides a second level of check to correct misclassifications by the graph model. By combining the Graph model with LLM-based reasoning, DISCOG enhances analysis accuracy, with the LLM acting as a validation and correction mechanism.

\section{Deployment and Business Impact}

DISCOG seamlessly integrates with existing eDiscovery solutions, significantly reducing the manual review workload. The heterogeneous graph can be constructed from similar databases on any system and stored on prem or in dedicated databases. The models used for prediction are light-weight and can be run on any infrastructure, with or without GPUs, while the LLM can be used from cloud services or open-source depending on the use-case and cost availability. Experiments on the ENRON dataset show that DISCOG achieves 80\% recall while requiring review of less than 10\% of the dataset. The approach scales efficiently to large eDiscovery datasets with minimal modifications, reducing false positives while maintaining low false negative rates.

According to market reviews in 2023, the document review process constitutes approximately 66\% of the total expenditure in the eDiscovery business\footnote{https://complexdiscovery.com/a-2022-look-at-ediscovery-processing-task-spend-and-cost-data-points/}, with the cost per document review ranging between \$0.50 to \$1.00, varying depending on the experience level of the reviewer and even higher for onsite reviews \footnote{https://edrm.net/2023/12/shaping-ediscovery-strategies-winter-2024-pricing-report/}. Leveraging DISCOG, deployable on-premise or on a low-cost cloud instance, significantly reduces costs by reducing the number of documents requiring manual review bringing down the overall cost to 10\%-20\% of the traditional process. For a database with millions of documents, DISCOG eliminates majority of the documents, thereby reducing the database size from millions to approximately to 10,000 - 20,000 documents, because of its high recall rate. This in turn reduces the overall review cost of the entire corpus to 1\%-2\% of its original cost, achieving approximately 98\% cost reduct. Further explanation  and calculations of the cost saving is added in Appendix \ref{Business_impact_appendix}.

\section{Conclusions}

We introduce \textbf{DISCOG}, a graph-based approach for predictive coding in eDiscovery, outperforming existing solutions in both classification and ranking tasks. Our analysis demonstrates its high accuracy, recall, and substantial cost savings compared to industry-standard methods.  


Future work will focus on benchmarking the system’s interpretability against manual reasoning and improving further scalabitility with open source LLMs for on-prem deployments. This hybrid approach aims to provide clear, interpretable justifications for the graph model's predictions, further improving the review process and fostering greater trust in automated document review systems.

\clearpage


\appendix

\section{Appendix}
\label{sec:appendix}
\subsection{Dataset}
\label{dataset_appendix}

 \begin{figure*}[t]
    \centering
    \pgfplotstableread{
x  srel snotrel qrelsrel qrelsnotrel
1 66 215 173 1303
2 203 175 122 1135
3 37 388 225 1302
4 44 437 387 1641
5 140 723 274 968
6 10 118 113 1318
7 69 172 806 1073
8 668 891 1268 1978
9 137 1307 411 2801
10 147 1233 249 3016
}{\mytable}

\begin{tikzpicture}
\begin{axis} [
    ybar,
    bar width=5pt,
    xticklabels={201, 202, 203, 204, 205, 206, 207, 401, 402, 403},
    xtick={1,2,3,4,5,6,7,8,9,10},
    xlabel={Topic},
    cycle list name = mycyclelist,
    width=16cm,
    height=4cm,
    xmin=0.5,
    xmax=10.5,
    ymin=0,
    ylabel={Number of Documents},
    ymode=log,
    legend columns=4,
    legend style={at={(0.8,-0.4)},anchor=north east},
    every axis plot/.append style={fill},
    no marks,
]
\addplot+ [] plot [] table [x=x, y=srel] {\mytable};
\addlegendentry{Seed Relevant}
\addplot+ [] plot [] table [x=x, y=snotrel] {\mytable};
\addlegendentry{Seed Not Relevant}
\addplot+ [] plot [] table [x=x, y=qrelsrel] {\mytable};
\addlegendentry{QRels Relevant}
\addplot+ [] plot [] table [x=x, y=qrelsnotrel] {\mytable};
\addlegendentry{QRels Not Relevant}
\end{axis} 
\end{tikzpicture}
    \caption{Distribution of relevant and non-relevant emails across the seed dataset and qrels for various topics. }
    \label{fig:image_label}
\end{figure*}

We primarily concentrate on production requests from the TREC Legal Tracks of 2009 and 2011, which include seven distinct topics for 2009, \textbf{Prepay Transactions (201)}, \textbf{FAS 140 (202)}, \textbf{Financial Forecasts (203)}, \textbf{Disposal of Documents (204)}, \textbf{Energy Loads (205)}, \textbf{Company's Financial Condition (206)}, and \textbf{Football Activities (207)} and three distinct topics for 2011, \textbf{Online Trading (401)}, \textbf{Derivative Trading (402)} , and \textbf{Environmental Impact (403)}. The distribution of the seed and qrels sets for each topic is shown in Fig. \ref{fig:image_label}.

\subsection{Hyper-parameter Tuning}

Hyper-parameter tuning was performed for all models, with a focus on optimizing epochs, learning rate, and batch size. For the KGE methods, the number of epochs ranged from 300 to 600 to achieve reasonable validation loss results. For GNN models, the number of epochs varied from 50 to 150 for GraphSAGE, and from 1000 to 2000 for GAT and RGCNs. Lower learning rates were applied for imbalanced data distributions, with fewer epochs for balanced datasets. The learning rate was adjusted within the range of 0.001 to 0.0001, while batch sizes varied from 128 to 1024 for GNN methods and around 100,000 for KGE methods. The hidden layer vector dimensions for GNNs were also tuned, with values ranging from 32 to 256.

\begin{figure*}[!th]
    \centering
    \pgfplotstableread{
x  NKNI NKYI YKNI YKYI 
1 0.75 0.82 0.84 0.9
2 0.75 0.8 0.74 0.89
3 0.38 0.71 0.85 0.87
4 0.89 0.63 0.85 0.77
5 0.51 0.91 0.91 0.92
}{\fscorefourzeroone}

\pgfplotstableread{
x  NKNI NKYI YKNI YKYI 
1 0.87 0.88 0.86 0.89
2 0.87 0.87 0.75 0.89
3 0.3 0.78 0.86 0.87
4 0.89 0.65 0.85 0.78
5 0.56 0.91 0.91 0.93
}{\precisionfourzeroone}

\pgfplotstableread{
x  NKNI NKYI YKNI YKYI 
1 0.74 0.8 0.83 0.9
2 0.74 0.78 0.76 0.89
3 0.5 0.7 0.85 0.87
4 0.89 0.66 0.86 0.79
5 0.53 0.91 0.91 0.91
}{\recallfourzeroone}

\pgfplotstableread{
x  NKNI NKYI YKNI YKYI 
1 0.71 0.75 0.74 0.82
2 0.61 0.74 0.66 0.78
3 0.5 0.53 0.62 0.63
4 0.88 0.51 0.84 0.54
5 0.81 0.89 0.89 0.89
}{\fscorefourzerotwo}

\pgfplotstableread{
x  NKNI NKYI YKNI YKYI 
1 0.95 0.9 0.86 0.81
2 0.81 0.88 0.64 0.74
3 0.5 0.53 0.64 0.62
4 0.91 0.52 0.83 0.54
5 0.79 0.92 0.91 0.89
}{\precisionfourzerotwo}

\pgfplotstableread{
x  NKNI NKYI YKNI YKYI 
1 0.65 0.69 0.69 0.82
2 0.58 0.69 0.69 0.84
3 0.5 0.53 0.8 0.67
4 0.86 0.51 0.85 0.54
5 0.84 0.86 0.87 0.89
}{\recallfourzerotwo}

\pgfplotstableread{
x  NKNI NKYI YKNI YKYI 
1 0.83 0.83 0.83 0.83
2 0.8 0.82 0.58 0.77
3 0.48 0.48 0.55 0.6
4 0.85 0.48 0.77 0.55
5 0.62 0.87 0.85 0.88
}{\fscorefourzerothree}

\pgfplotstableread{
x  NKNI NKYI YKNI YKYI 
1 0.97 0.97 0.97 0.96
2 0.97 0.9 0.57 0.81
3 0.46 0.46 0.58 0.58
4 0.84 0.46 0.77 0.55
5 0.62 0.86 0.85 0.88
}{\precisionfourzerothree}

\pgfplotstableread{
x  NKNI NKYI YKNI YKYI 
1 0.76 0.76 0.76 0.76
2 0.73 0.76 0.6 0.74
3 0.5 0.5 0.54 0.66
4 0.86 0.5 0.76 0.59
5 0.63 0.89 0.85 0.89
}{\recallfourzerothree}

\pgfplotstableread{
x  NKNI NKYI YKNI YKYI 
1 0.76 0.8 0.8 0.85
2 0.72 0.79 0.66 0.81
3 0.45 0.57 0.67 0.7
4 0.87 0.54 0.82 0.62
5 0.65 0.89 0.88 0.9
}{\fscoreaverage}

\pgfplotstableread{
x  NKNI NKYI YKNI YKYI 
1 0.93 0.92 0.9 0.89
2 0.88 0.88 0.65 0.81
3 0.42 0.59 0.69 0.69
4 0.88 0.54 0.82 0.62
5 0.66 0.9 0.89 0.9
}{\precisionaverage}

\pgfplotstableread{
x  NKNI NKYI YKNI YKYI 
1 0.72 0.75 0.76 0.83
2 0.68 0.74 0.68 0.82
3 0.5 0.58 0.73 0.73
4 0.87 0.56 0.82 0.64
5 0.67 0.89 0.88 0.9
}{\recallaverage}

\begin{tikzpicture}
    \begin{groupplot}[
        group style={group size=3 by 3,
            horizontal sep = .2 cm, 
            vertical sep = .2 cm,            
            ylabels at=edge left,
            xlabels at=edge bottom,
            xticklabels at=edge bottom,
            yticklabels at=edge left}, 
        xlabel = {},
        xticklabel style={align=center},
        xticklabels = {TransE, ComplEx, GAT, RCGN\\(TransE), Graph\\SAGE},
        xtick={1,2,3,4,5},
        ymin=0,
        ymax=1.1,
        ybar=1pt,
        /pgf/bar width=1.5pt,
        ylabel = {},
        width = 6.3 cm, 
        height = 3 cm,
        legend columns=2, 
        legend style={at={(8,-4.0)},anchor=north},
        cycle list name = mycyclelist,
        every axis plot/.append style={fill},
        no marks,
        ]
        \nextgroupplot[title={\textbf{Online Trading}}, ylabel={F1 Score}]
        \addplot+ table [x=x, y=NKNI] {\fscorefourzeroone};        
        \addplot+ table [x=x, y=NKYI] {\fscorefourzeroone};        
        \addplot+ table [x=x, y=YKNI] {\fscorefourzeroone};
        \addplot+ table [x=x, y=YKYI] {\fscorefourzeroone};

        \nextgroupplot[title={\textbf{Derivative Trading}}]
        \addplot+ table [x=x, y=NKNI] {\fscorefourzerotwo};        
        \addplot+ table [x=x, y=NKYI] {\fscorefourzerotwo};        
        \addplot+ table [x=x, y=YKNI] {\fscorefourzerotwo};
        \addplot+ table [x=x, y=YKYI] {\fscorefourzerotwo};
        
        \nextgroupplot[title={\textbf{Environmental Impact}}]
        \addplot+ table [x=x, y=NKNI] {\fscorefourzerothree};        
        \addplot+ table [x=x, y=NKYI] {\fscorefourzerothree};        
        \addplot+ table [x=x, y=YKNI] {\fscorefourzerothree};
        \addplot+ table [x=x, y=YKYI] {\fscorefourzerothree};

        
        \nextgroupplot[ylabel={Precision},]
        \addplot+ table [x=x, y=NKNI] {\precisionfourzeroone};        
        \addplot+ table [x=x, y=NKYI] {\precisionfourzeroone};        
        \addplot+ table [x=x, y=YKNI] {\precisionfourzeroone};
        \addplot+ table [x=x, y=YKYI] {\precisionfourzeroone};
        
        \nextgroupplot[]
        \addplot+ table [x=x, y=NKNI] {\precisionfourzerotwo};        
        \addplot+ table [x=x, y=NKYI] {\precisionfourzerotwo};        
        \addplot+ table [x=x, y=YKNI] {\precisionfourzerotwo};
        \addplot+ table [x=x, y=YKYI] {\precisionfourzerotwo};
        
        \nextgroupplot[]
        \addplot+ table [x=x, y=NKNI] {\precisionfourzerothree};        
        \addplot+ table [x=x, y=NKYI] {\precisionfourzerothree};        
        \addplot+ table [x=x, y=YKNI] {\precisionfourzerothree};
        \addplot+ table [x=x, y=YKYI] {\precisionfourzerothree};

        
        \nextgroupplot[ylabel={Recall},]
        \addplot+ table [x=x, y=NKNI] {\recallfourzeroone};        
        \addplot+ table [x=x, y=NKYI] {\recallfourzeroone};        
        \addplot+ table [x=x, y=YKNI] {\recallfourzeroone};
        \addplot+ table [x=x, y=YKYI] {\recallfourzeroone};
        
        \nextgroupplot[single ybar legend,legend to name=zelda, xlabel={},x label style={at={(-0.5,-0.03)}}]
        \addplot+ table [x=x, y=NKNI] {\recallfourzerotwo};  
        \addlegendentry{Keywords - No, Sender/Reciever Ids - No\quad};
        \addplot+ table [x=x, y=NKYI] {\recallfourzerotwo};   
        \addlegendentry{Keywords - No, Sender/Reciever Ids - Yes\quad};
        \addplot+ table [x=x, y=YKNI] {\recallfourzerotwo};
        \addlegendentry{Keywords - Yes, Sender/Reciever Ids - No\quad};
        \addplot+ table [x=x, y=YKYI] {\recallfourzerotwo};
        \addlegendentry{Keywords - Yes, Sender/Reciever Ids - Yes\quad};
        
        \nextgroupplot[]
        \addplot+ table [x=x, y=NKNI] {\recallfourzerothree};        
        \addplot+ table [x=x, y=NKYI] {\recallfourzerothree};        
        \addplot+ table [x=x, y=YKNI] {\recallfourzerothree};
        \addplot+ table [x=x, y=YKYI] {\recallfourzerothree};

    \end{groupplot}   
    \ref{zelda}
\end{tikzpicture}
    \caption{Ablation study of attributes added to the graph in the form of nodes.}
    \label{fig:ablation}
\end{figure*}

\subsection{Ablation Study} 
\label{appendix_ablation}

To assess the impact of different graph attributes, we conduct an ablation study that systematically evaluates the necessity of various node types within the graph. Given space constraints, this study focuses on three representative topics (401, 402, and 403, which are described as Online Trading, Derivative Trading, and Environmental Trading respectively). These topics capture a range of relevant and non-relevant distributions. However, the DISCOG methodology can be extended to other topics discussed in this paper.

In this ablation study, the base graph structure consists of two core node types: Emails and Topics, which remain constant across all experiments. To explore the effect of additional features, we incrementally add different combinations of nodes---specifically, keyword nodes and sender/receiver nodes. The influence of these additions is assessed by analyzing their impact on the predictive coding results, measured against the \emph{qrels} set. Importantly, the model architecture and training hyperparameters are held constant across all experiments to ensure that observed differences are solely due to the variations in the graph structure.

The results, as shown in Fig.~\ref{fig:ablation}, reveal that incorporating keyword nodes and sender/receiver nodes, along with establishing similarity links between them, leads to a marked improvement in the overall performance metrics of the models. These improvements are consistent across several models, with the exception of the RGCN model, which shows little to no performance gain from the additional graph attributes. This suggests that the effectiveness of graph augmentation may depend on the underlying model architecture, with some models being more sensitive to additional structural information than others.

\subsection{Business Impact Calculations}
\label{Business_impact_appendix}

Assuming review cost per document is \$0.5 - \$1.0, depending on the type of review, the cost of reviewing a million documents for any eDiscovery use case ranges between \$500,000 to \$1,000,000. With the use of DISCOG, the number of documents tagged for review is reduced to 10\% -20\% of the original corpus, which is approximately 10,000 - 20,000 documents from the entire corpus (assuming a cutoff of 20,000 documents requiring manual review). This cutoff is determined based on the Recall@k metrics, where the DISCOG method with GraphSAGE algorithm achieves over 80\% recall. By analyzing the cost of \$0.50 to \$1.00 per document for 20,000 documents instead of the entire corpus, the cost for the entire corpus is reduced to \$10,000 - \$20,000, which translates to a per document cost of \$0.01 to \$0.02 on average for the entire corpus. Consequently, our method requires only 1\% to 2\% of the manual review cost.

\end{document}